\documentclass[conference]{IEEEtran}
\IEEEoverridecommandlockouts
\usepackage{cite}
\usepackage{amsmath,amssymb,amsfonts}
\usepackage{algorithmic}
\usepackage{graphicx}
\usepackage{textcomp}
\usepackage{xcolor}

\usepackage{booktabs}
\usepackage{adjustbox}
\usepackage{multirow}
\usepackage{multicol}
\usepackage{enumitem}
\usepackage{bm} 
\usepackage{booktabs}
\usepackage{subfigure}
\usepackage{enumitem}

\def\BibTeX{{\rm B\kern-.05em{\sc i\kern-.025em b}\kern-.08em
    T\kern-.1667em\lower.7ex\hbox{E}\kern-.125emX}}
\begin{document}

\title{$x$\textit{Time}: Extreme Event Prediction with Hierarchical Knowledge Distillation and Expert Fusion}

\author{\IEEEauthorblockN{Quan Li$^{1}$, Wenchao Yu$^{2}$, Suhang Wang$^{1}$, Minhua Lin$^{1}$, Lingwei Chen$^{3}$, Wei Cheng$^{2}$, Haifeng Chen$^{2}$}
\IEEEauthorblockA{
$^1$\textit{Pennsylvania State University},
University Park, PA, USA \\
$^2$\textit{NEC Labs America}, New Jersey, NJ, USA\\
$^3$\textit{Rochester Institute of Technology}, Rochester, NY, USA\\
\{qbl5082, szw494, mfl5681\}@psu.edu, 
\{wyu, weicheng, haifeng\}@nec-labs.com, lwcics@rit.edu}
}

\maketitle

\begin{abstract}
Extreme events frequently occur in real-world time series and often carry significant practical implications. In domains such as climate and healthcare, these events, such as floods, heatwaves, or acute medical episodes, can lead to serious consequences. Accurate forecasting of such events is therefore of substantial importance. Most existing time series forecasting models are optimized for overall performance within the prediction window, but often struggle to accurately predict extreme events, such as high temperatures or heart rate spikes. The main challenges are data imbalance and the neglect of valuable information contained in intermediate events that precede extreme events. In this paper, we propose xTime, a novel framework for extreme event forecasting in time series. xTime leverages knowledge distillation to transfer information from models trained on lower-rarity events, thereby improving prediction performance on rarer ones. In addition, we introduce a mixture of experts (MoE) mechanism that dynamically selects and fuses outputs from expert models across different rarity levels, which further improves the forecasting performance for extreme events. Experiments on multiple datasets show that xTime achieves consistent improvements, with forecasting accuracy on extreme events improving from 3$\%$ to 78$\%$.
\end{abstract}


\section{Introduction}
Time series forecasting plays a fundamental role across a broad spectrum of critical applications, such as stock market analysis, weather and climate modeling, and electricity demand prediction. Various time series forecasting approaches have been developed. Traditional time series forecasting approaches~\cite{box2015time,sapankevych2009time} 
have long dominated the field, relying on assumptions like linearity and stationarity. 
Meanwhile, models based on neural networks like recurrent neural networks~\cite{hewamalage2021recurrent} and Transformers~\cite{Yuqietal-2023-PatchTST} have become increasingly popular and significantly improved predictive performance by learning richer temporal patterns. Recently, researchers have started to explore the application of large language models (LLMs) for time series forecasting, either by adapting the underlying LLM-based architectures to temporal data or by directly training LLMs on large-scale time series datasets~\cite{ansari2024chronos,shi2024time,jin2023timellm,wang2024news,lin2024decoding}. 

    

Despite these advancements, most existing time series forecasting methods focus on improving overall performance on all points within the prediction window, which inherently biases them toward the frequent patterns found in normal events, making them struggle to detect and forecast extreme events; while the prediction of extreme events, which are rare but significant deviations from normal behavior, is of critical importance. Extreme events, despite their infrequency, often act as early indicators of high-impact disruptions with serious consequences~\cite{omar2022exploring}. As such, accurately predicting extreme events is essential in domains where timely alerts and proactive risk mitigation strategies are crucial for preventing cascading failures or minimizing losses. The key challenges in extreme event forecasting are data imbalance~\cite{he2021weighting} and the neglect of valuable information contained in intermediate events that precede extreme events \cite{faletto2023predicting}, which is largely overlooked in mainstream approaches. In practice, the rarer an event is, the fewer training samples are available, making it difficult for models to learn meaningful and representative patterns. As a result, models tend to underpredict extreme events, and their prediction performance deteriorates as event rarity increases.


\begin{figure*}[t]
    \centering
    \includegraphics[width=0.9\linewidth]{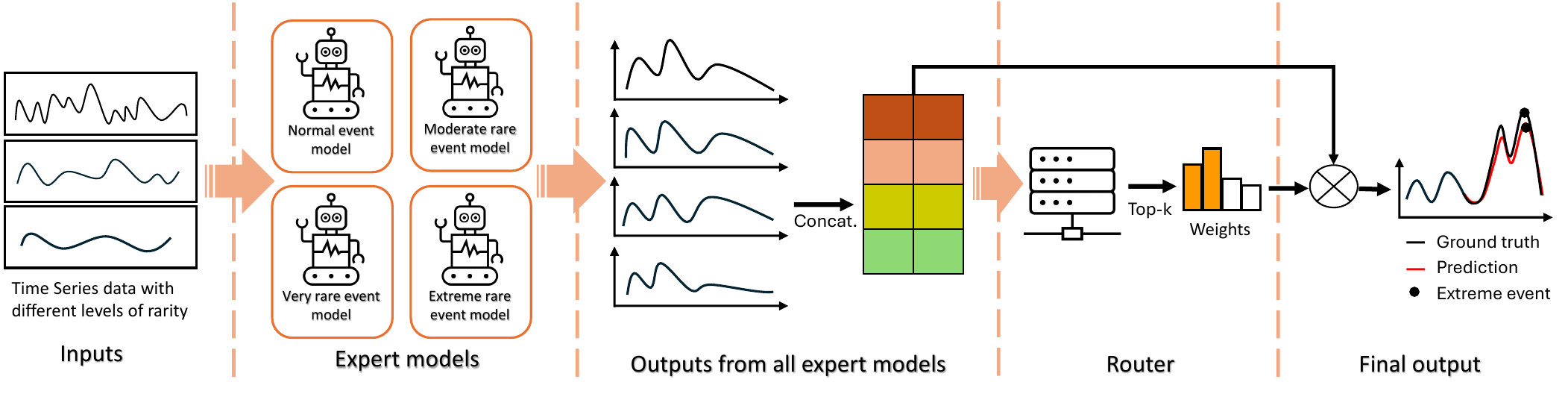}
    \vskip -1em
    \caption{The workflow of the proposed xTime framework: the input time series is first processed by pretrained expert models specialized for different rarity levels, and the router then fuses their outputs to generate the final prediction.}
    \label{fig:overview}
    \vskip -1.5em
\end{figure*}

In recent years, the importance of accurately forecasting extreme events in time series data has garnered growing attention from the research community. Several works have tackled this problem by using extreme value theory (EVT) or designing specialized loss functions tailored to better capture the statistical properties of rare events \cite{ding2019modeling,wilson2022deepgpd,Liu2023Sadi,wang2024self}. Other approaches have adopted classifier-based pipelines, where models first identify the presence of extreme events and then predict their future trajectory or impact~\cite{hou2021deep,li2023extreme}, or focus exclusively on block maxima within sliding prediction windows to isolate and forecast rare occurrences \cite{galib2023self}. These methods have demonstrated the improvements in forecasting extreme events and partially mitigated data imbalance through various strategies. 
However, these methods exhibit certain limitations. For instance, classifier-based methods are fragile, as their accuracy heavily depends on the initial classification stage, making them difficult to deploy in practice. Moreover, these approaches overlook the potential relationships between extreme events and intermediate events.

In real-world systems, extreme events do not occur in isolation but rather emerge as the cumulative outcome of a sequence of lower-rarity or precursor events~\cite{faletto2023predicting}. For example, in climate systems, a heatwave is often preceded by a sequence of gradually intensifying warm days. Similarly, in healthcare, heart rate spikes are frequently preceded by a series of tachycardia episodes or arrhythmias serving as warning signs. This observation underscores the need for forecasting models that not only specialize in rare event detection but also exploit relationships across different rarity levels to improve forecasting fidelity and robustness. 

In this paper, we propose a novel extreme event forecasting framework, xTime,
that leverages knowledge distillation to transfer knowledge from common intermediate events and lower-rarity expert models, employs multi-resolution analysis to capture temporal patterns at different scales, and integrates the MoE paradigm to learn the ranking relationships among rarity-specific expert models. Our approach assumes that time series events can be categorized into different rarity levels based on their frequency, which can be flexibly defined according to the application needs. For example, events that occur in less than $1\%$, between $1\%$ and $5\%$, and between $5\%$ and $10\%$ of the data distribution can be defined as extreme events, very rare events, and moderately rare events, respectively~\cite{shyalika2024comprehensive}.

As the high-level workflow of xTime shown in Fig~\ref{fig:overview}, the input time series data are firstly processed with the dedicated expert models that are trained for predefined rarity levels by using only the data corresponding to the rarity levels. To overcome data imbalance and utilize the relationship between consecutive rarity levels, we introduce a hierarchical knowledge distillation strategy, where lower-rarity expert models act as teacher models to guide the training of higher-rarity student models. This enables rare event models to benefit from transferable patterns learned from more frequent events, improving generalization in data-scarce settings. To further enhance specialization and maintain inter-level consistency, a carefully designed rare penalty loss is introduced to encourage each expert to focus more effectively on its designated rare level while still capturing the underlying relationships with neighboring levels. Additionally, we apply the wavelet transform to decompose the input time series into multi-resolution components in the time-frequency domain, allowing each expert to analyze data from diverse perspectives. Then, inspired by the MoE paradigm, we design a router that takes the outputs of all expert models as input and dynamically selects and assigns weights to them based on their relative ranking. The final forecasting output that includes extreme events is obtained by fusing the predictions of the expert models based on the assigned weights, enabling adaptive integration of knowledge across different rarity levels.

The main contributions of this work are as follows:
\begin{itemize}
    \item We propose a knowledge distillation strategy to bridge the gap between common, intermediate, and extreme events across rarity levels. By transferring knowledge from lower-rarity models to those targeting higher-rarity events, our method mitigates data imbalance and improves extreme event forecasting.

    \item We introduce a novel expert fusion mechanism inspired by the MoE paradigm. By leveraging the ordinal relationship among rarity levels, we design a router to integrate predictions from expert models, effectively utilizing information from intermediate events to support extreme event forecasting.

    \item We design a rarity-aware penalty loss that encodes the ranking structure of rarity levels. This loss encourages each expert model to concentrate on its corresponding level of rarity while respecting inter-level relationships.

    \item Extensive experiments on real-world time series datasets demonstrate the effectiveness and robustness of xTime, significantly outperforming existing methods on extreme event prediction across all rarity levels.
\end{itemize}

\section{Related Works}
\subsection{Time Series Forecasting}

Time series prediction is a fundamental task in time series research. Early studies often relied on linear models such as autoregressive moving averages \cite{said1984testing, box2015time}, or nonlinear approaches like the Nonlinear Autoregressive with Exogenous (NARX) model \cite{lin1996learning}. However, the effectiveness of these traditional methods is limited by their shallow architectures and poor ability to generalize across diverse time series patterns.

To overcome these limitations, with the development of deep learning, recent advancements have turned to deep learning techniques, which offer stronger representation learning capabilities. Models such as Recurrent Neural Networks (RNNs) \cite{sherstinsky2020fundamentals}, Long Short-Term Memory (LSTM) networks \cite{hochreiter1997long}, and Gated Recurrent Units (GRUs) \cite{cho2014learning} have shown significant improvements in capturing temporal dependencies in sequential data. More recently, Transformer-based architectures \cite{Yuqietal-2023-PatchTST,wu2021autoformer,zhou2021informer} have gained popularity in time series forecasting due to their ability to model long-range dependencies and parallelize computation. These developments have led to state-of-the-art performance across a variety of time series tasks, particularly in scenarios involving complex patterns or extreme events.
Although these proposed methods have achieved success in general time series forecasting tasks, they do not explicitly tackle the challenge of extreme event prediction, which is often hindered by inherent data imbalance in the datasets.

\subsection{Large Time Series Foundation Models}

Recently, with the rapid development and success of LLMs, researchers have begun exploring their application in the time series domain. This emerging line of work can be broadly categorized into two directions: (1) leveraging existing pre-trained LLMs directly for time series tasks, and (2) training large-scale time series foundation models from scratch.


To leverage existing pre-trained LLMs, several approaches have been proposed. Time-LLM \cite{jin2023timellm} employs a reprogramming technique to adapt the objective of LLMs from text generation to time series forecasting. CALF \cite{liu2025calf} and TimeCMA \cite{liu2024timecma} adopt multi-modal strategies by aligning time series data with textual information through different alignment mechanisms to improve predictive performance. ChatTime \cite{wang2025chattime} introduces uniformly distributed numerical tokens into the vocabulary to fine-tune LLMs specifically for time series tasks.
On the other hand, several large-scale time series foundation models have been developed from scratch. Among them, Moirai \cite{woo2024moirai}, Chronos \cite{ansari2024chronos}, Time-MoE \cite{shi2024time}, and TimesFM \cite{das2024decoder} are some of the most prominent. These models leverage various LLM-inspired architectures and are trained on large-scale datasets to capture complex temporal dependencies and perform accurate time series forecasting. Benefiting from large-scale training data, these LLM-based time series models outperform traditional methods on general forecasting tasks. However, their effectiveness in predicting rare extreme events remains limited due to the inherent data imbalance, where most training samples represent normal patterns, leaving rare events underrepresented during learning. 

\subsection{Extreme Event Prediction}

In recent years, as time series research has advanced, increasing attention has been given to extreme events due to their critical practical significance. For instance, extreme weather conditions, such as heatwaves or severe cold spells, pose serious risks and require timely prediction. However, because these events are rare by nature, accurately forecasting them remains a significant challenge.

Ding et al. \cite{ding2019modeling} proposed EVL, which leverages Extreme Value Theory to model extreme events within datasets and build predictive models specifically for these rare occurrences. NEC+ \cite{li2023extreme} trains three separate models (normal model, extreme model, and classifier) to improve extreme event prediction. The classifier distinguishes between normal and extreme inputs, directing data to the corresponding prediction model. SaDI \cite{said1984testing} is a feature-engineering-based approach that utilizes specific features to enhance extreme event forecasting. EPL \cite{wang2024self} introduces an extreme penalty loss designed to better capture extreme events during training. ReFine \cite{shi2024refine} improves prediction by applying reweighting and refinement techniques focused on extreme data. However, existing extreme event prediction methods largely overlook the relationships between common, intermediate, and extreme events. As highlighted by Faletto et al. \cite{faletto2023predicting}, extreme events often arise as the cumulative outcome of more frequent and intermediate occurrences. Current approaches tend to either focus solely on extreme event prediction or train models for normal and extreme events separately. Additionally, Galib et al. \cite{galib2023self} pointed out that extreme events typically correspond to block maxima within time series data. Apart from EPL, most existing methods concentrate exclusively on the raw time series data and neglect the potential benefits of using decomposed time series components.

Unlike existing extreme event prediction approaches, we utilize decomposed time series data to perform a more in-depth analysis from multiple perspectives. Additionally, we employ knowledge distillation \cite{li2024enhancing,li2022distilling} to transfer knowledge from intermediate events, thereby enhancing the prediction of extreme events.

\section{Problem Statement}
    In this work, we study the problem of extreme event prediction in time series forecasting. Given a univariate time series $\mathbf{X} \in \mathbb{R}^T = \{x_0, x_1, \dots, x_t\}$ over a historical window of length $T$, the objective is to forecast the future values $\mathbf{Y}_{T:T+H} = \{y^c_t, y^c_{t+1}, \dots, y^c_{t+h}\}$ over a prediction horizon $H$, where $c$ denotes the rarity levels. We aim to minimize the error for $y^c$ with $c \in \mathcal{C}_{rare}$ within the prediction window. Formally, we define the extreme event prediction task as:
    \begin{equation}
        Y_{T:T+H} = \underset{y^c \in \mathbf{Y}_{T:T+H} | c \in \mathcal{C}_{rare}}{\arg\!\min} f(X_{1:T}, \{M\}_{e=1}^E, \mathcal{R})
    \end{equation}
    where $\mathcal{C}_{rare}$ is the target rarity levels, ${M}_{e=1}^E$ denotes a set of expert models, each tailored to a specific rarity level and trained with knowledge distillation, and $\mathcal{R}$ is a router that dynamically selects and combines expert outputs.

    To define rarity levels, we follow the categorization from \cite{shyalika2024comprehensive}, which divides events into four levels based on their frequency in the data distribution. Formally, the $i$-th data point $y_i$ is assigned a rarity level $c \in \mathcal{C} = \{normal, moderate\ rare, very\ rare, extreme\ rare\}$ based on the following: $c = extreme\ rare$ if $y_i > P_{99}$, $c = very\ rare$ if $P_{95} < y_i \leq P_{99}$, $c = moderate\ rare$ if $P_{90} < y_i \leq P_{95}$, and $c = normal$ if $y_i \leq P_{90}$, 
    where $P_k$ denotes the $k$-th percentile of the data distribution. In other words, if the assigned rarity level $c$ for a point $y_i$ is $extreme \ rare$, then $y_i$ is defined as an extreme event; other levels follow similarly. In time series prediction tasks, where the prediction length typically spans multiple time steps, we define the rarity level of a sequence based on the maximum value in the prediction window. For example, if the maximum value in a sequence falls within the extreme range, the whole sequence is labeled as extreme. 
    Note that the thresholds for rarity levels are adaptable and can be defined flexibly based on the requirements of specific applications and datasets.

\section{Proposed Method}

    In this section, we present the details of the proposed method, xTime. An overview of the xTime workflow is illustrated in Fig. \ref{fig:overview}. The input time series is first processed by multiple expert models, and a router dynamically assigns weights to each expert. The weighted outputs are then aggregated to produce the final forecasting result. Each expert model is pretrained using knowledge distillation and trained with a hierarchical rare penalty loss to enhance its specialization. Meanwhile, the router is pretrained using a pointwise ranking loss to effectively guide expert selection. The following subsections elaborate on each component in detail.

\subsection{Expert Models}
    \begin{figure}
        \centering
        \subfigure[Expert model training]{\includegraphics[width = 0.8\linewidth]{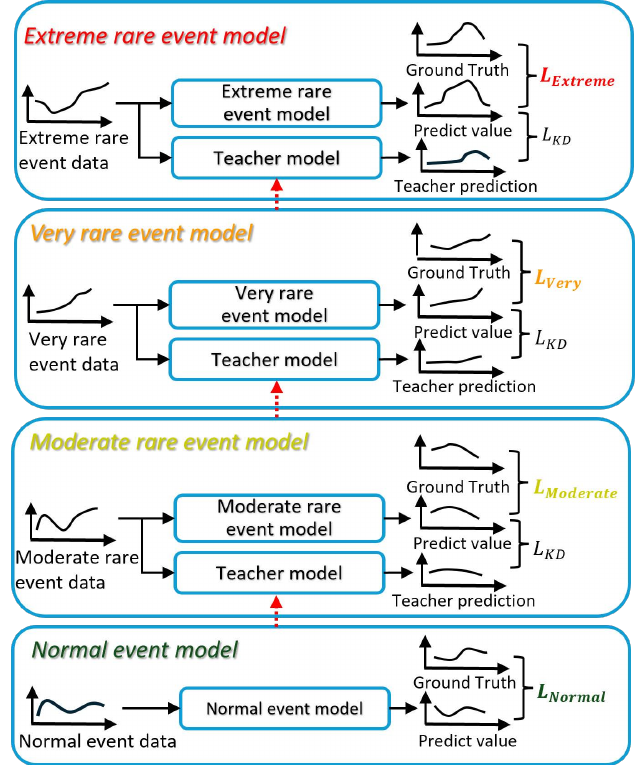} \label{subfig:expertmodeltraining}} 
        \subfigure[Expert model architecture]{\includegraphics[width = 0.9\linewidth]{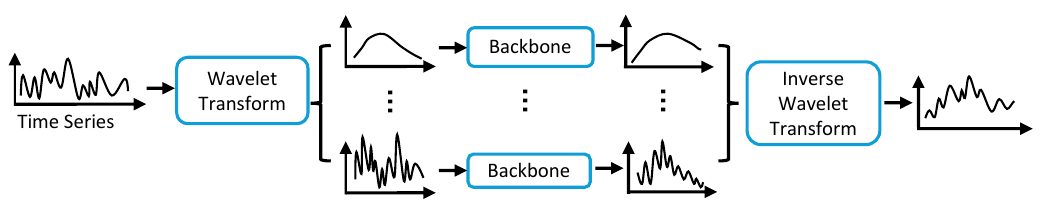} \label{subfig:expertmodelarch}} 
        
        \caption{a) The training mechanism for rarity expert models and b) The detailed architecture for all expert models}
        \vskip -1em
        \label{fig:expertmodel}
    \end{figure}


The training process for expert models and the detailed architectures are shown in Fig~\ref{fig:expertmodel}. In this subsection, we first present the detailed architecture of all expert models, then describe the training strategy that facilitates knowledge transfer from intermediate to rare events, and introduce the rarity-aware loss function specifically designed to guide each expert model toward its target rarity level. Note that Fig.~\ref{fig:expertmodel} show 4 expert models as an example; the actual number of expert models is flexible based on application requirements.

\subsubsection{Model Architecture}
As shown in Fig~\ref{subfig:expertmodelarch}, to better capture the underlying structure and dynamic variations in time series, we employ the Empirical Wavelet Transform (EWT) \cite{gilles2013empirical} to decompose the input signal. Unlike traditional methods such as Fast Fourier Transform (FFT) \cite{nussbaumer1982fast}, which operates purely in the frequency domain, or moving average (MA) methods that decompose signals in the time domain into trend and seasonal components, EWT offers a more flexible decomposition by adapting to both time and frequency characteristics of the signal.

    EWT constructs adaptive wavelet filters based on the empirical spectrum of the signal, allowing it to capture meaningful frequency components that vary over time and enabling a joint time-frequency representation. Compared to Short-Time Fourier Transform (STFT) \cite{griffin1984signal} or classical wavelet transforms, EWT does not require predefining fixed windows or basis functions. Instead, it automatically determines frequency boundaries and constructs wavelet filters accordingly.

    As extreme events in time series often manifest as block maxima or abrupt fluctuations over short durations \cite{galib2024fide}, wavelet-based decomposition, which combines the benefits from both temporal and frequency domains and is sensitive to the mutation signals, is particularly well-suited for analyzing and forecasting such events.

    Formally, given $\mathbf{X} \in \mathbb{R}^T$, the EWT decomposes $\mathbf{X}$ into $B$ components as 
    $\mathbf{X} = \sum_{b=1}^{B} \mathbf{X}_b$,
   where $B$ denotes the number of frequency bands. Each $\mathbf{X}_b$ captures the information in $b$-th frequency band, enabling fine-grained multi-resolution analysis.
    
    To obtain $\mathbf{X}_b$, EWT first computes its Fourier spectrum $\hat{\mathbf{X}}(\omega) = FFT(\mathbf{X})$.
    Then a set of frequency boundaries $\{ \omega_b \}_{b=0}^B$ are automatically determined by detecting local maxima or energy variations in the Fourier spectrum $|\hat{\mathbf{X}}(\omega)|$, which can be described as 
    $[0, \pi]=\bigcup_{b=1}^{B} [\omega_{b-1}, \omega_b]$, where $\omega_0 = 0$ and $ \omega_B = \pi$.
    For each frequency band $[\omega_{b-1}, \omega_b]$, a corresponding empirical wavelet $\hat{\psi}_b(\omega)$ is constructed as a smooth bandpass filter using trigonometric transition functions near the boundaries.
    The $\mathbf{X}_b$ is then obtained via inverse Fourier transform as 
    $\mathbf{X}_b = \mathcal{F}^{-1} \left[ \hat{\mathbf{X}}(\omega) \cdot \hat{\psi}_b(\omega) \right]$,
    where $\mathcal{F}^{-1}$ denotes the inverse Fourier transform. 
    
    EWT analyzes signal characteristics in the frequency domain while preserving temporal information through decomposition in the time domain. By inheriting the benefits of traditional wavelet transforms without requiring manual selection of wavelet bases, EWT enables xTime to effectively capture sharp transitions, rare fluctuations, and multi-scale patterns. This results in a richer, more informative representation for downstream extreme event prediction.

   We then employ a transformer-based model (e.g., PatchTST \cite{Yuqietal-2023-PatchTST}) as the backbone model to process each decomposed time series component obtained from EWT, leveraging the transformer’s capability to model complex dependencies and capture relationships between patterns.
    For each decomposed component $\mathbf{X}_b$, this process can be formulated as $\mathbf{y}_b= Transformer(\mathbf{X}_b)$,
    where $\mathbf{y}_b$ is the prediction corresponding to the decomposed signal $\mathbf{X}_b$. To reconstruct a unified prediction from these multi-resolution results, we apply the inverse EWT as $y^{expert} = \sum_{b=1}^{B} \mathbf{y}_b$.
    
    Note that our framework is model-agnostic. The backbone can be replaced with any state-of-the-art universal time series forecasting model without changing the overall structure.

\subsubsection{Expert Model Training With Knowledge Distillation}
    As suggested by prior studies, extreme events are often the cumulative outcome of more common intermediate events \cite{faletto2023predicting}. This indicates that high-rarity events can benefit from the patterns embedded in lower-rarity events. In other words, although extreme events suffer from severe data imbalance due to their rarity, the latent knowledge captured from more frequent, intermediate events offers an opportunity to mitigate this challenge. Motivated by this, we adopt a knowledge distillation strategy to transfer knowledge from expert models trained on lower-rarity events to those responsible for predicting higher-rarity events.

    As illustrated in Fig. \ref{subfig:expertmodeltraining}, each teacher model is trained on samples from a lower rarity level and shares the same architecture as its corresponding student model, which targets a higher rarity level. For instance, the model trained on very rare events serves as the teacher for the expert model handling extremely rare events. During training, we employ a knowledge distillation loss to guide the student model in learning from the teacher's predictions. This design enables the transfer of informative patterns from intermediate to rarer events, thereby establishing a connection across rarity levels and enhancing performance on data-scarce extreme events. 
    
    To facilitate effective knowledge transfer while allowing flexibility in extreme regions, we further introduce an adaptive temperature mechanism into the distillation process. Instead of using a fixed temperature, we assign higher temperatures to samples with larger prediction errors, which typically correspond to rare or extreme events. This design allows the student model to diverge from the teacher in cases where the teacher's prediction is unreliable, and instead focus more on learning directly from the ground-truth signal. Specifically, let $y_i^{student}$ and $y_i^{teacher}$ denote the outputs of the student and teacher models for the $i$-th data point, respectively. The knowledge distillation loss can be formulated as:
    \begin{equation}
        \small
        L_{KD} = \frac{1}{N}\sum_{i=1}^{N}(\frac{\Delta y_i}{T_i})^2, \Delta y_i = y_i^{student}-y_i^{teacher}, T_i = 1 + |\Delta y_i|
    \end{equation}
    where $N$ is the number of samples, and $T_i$ is the adaptive temperature for the $i$-th sample, dynamically adjusted based on the teacher's prediction error. This mechanism enhances the student's ability to learn fine-grained distinctions in high-rarity regions without being overly constrained by the teacher. Note that for the expert model of the lowest rarity level (i.e., normal events), we do not incorporate a teacher model, as we assume there is sufficient data, and the events primarily represent normal cases.

\begin{figure}
    \centering
    \vskip -1em
    \includegraphics[width=0.8\linewidth]{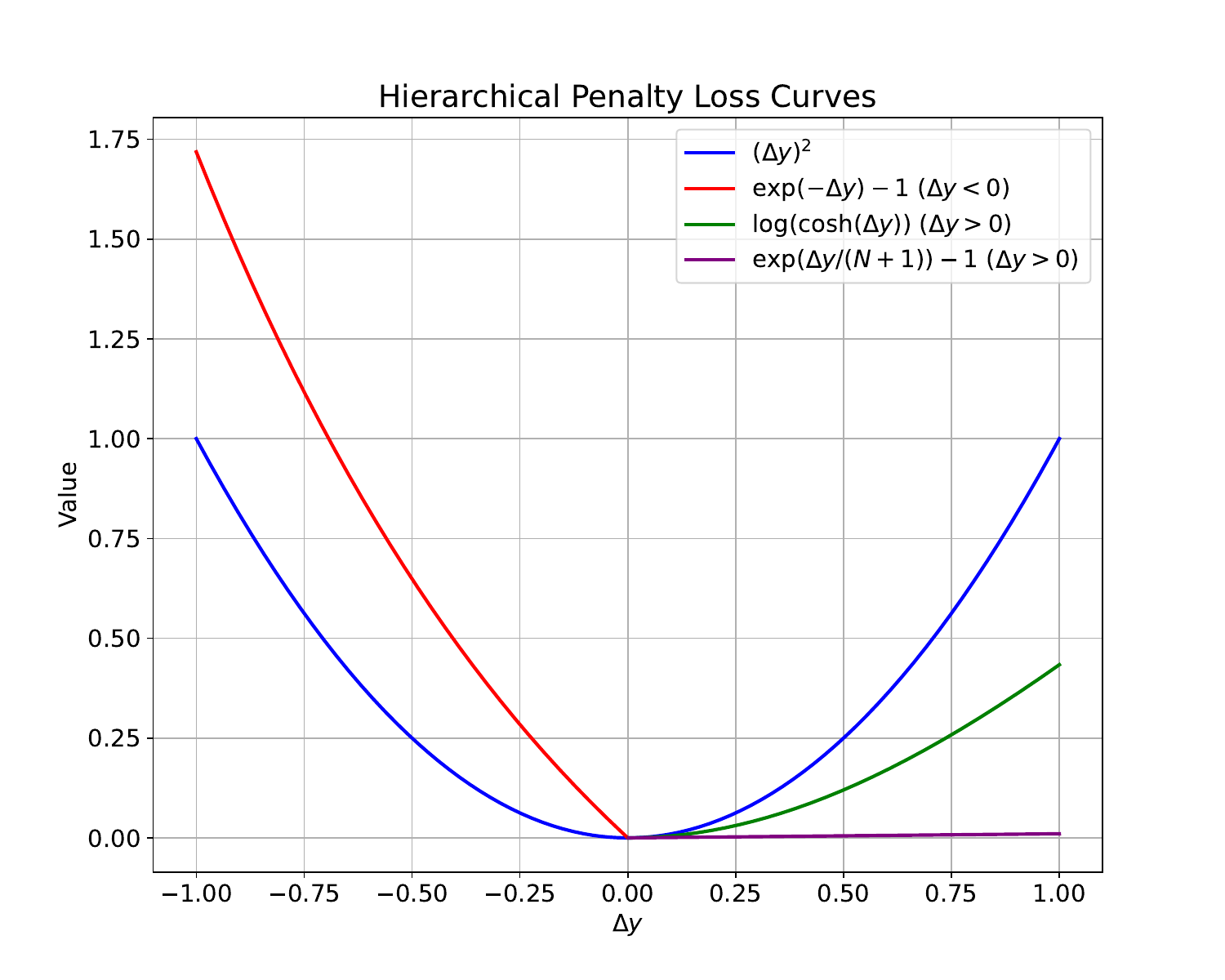}
    \vskip -1.5em
    \caption{Hierarchical rare penalty loss visualization}
    \vskip -1em
    \label{fig:h_loss}
        \vskip -1em
\end{figure}
    
\subsubsection{Hierarchical Rare Penalty Loss}

    To further guide each expert model toward accurately forecasting events within its designated rarity level, inspired by \cite{wang2024self}, we propose a hierarchical rare penalty loss that leverages the ranking relationships among rarity levels. Unlike prior loss-based approaches that dichotomize events into either normal or extreme, our formulation extends to a multi-rarity setting, allowing each expert to specialize in learning its assigned level.

    In our design, under-predictions of rare events incur stronger penalties, highlighting the importance of correctly recognizing these impactful but infrequent occurrences. For over-prediction, to address the scarcity of rare event data, we intentionally reduce the penalty. The penalty is adjusted based on the rarity level: the higher the rarity, the lower the penalty, encouraging the model to be more assertive in detecting rare cases. It reflects the scarcity of high-rarity samples and encourages each expert model to focus on its designated rarity level, preventing excessive over-prediction toward rarer events.

    Specifically, given the prediction result $Y^{student}=\{y_0^s, y_1^s, \dots,y_n^s\}$ and ground truth $Y^{ground\_truth}=\{y_0^{gt}, y_1^{gt}, \dots, y_n^{gt}\}$, the overall loss function can be formulated as follows, with the specific loss terms varying across different rarity levels:
    \begin{equation}
        L_{rare} =  \frac{1}{N} \sum_{i=1}^Nf(\Delta y_i),\Delta y_i=y_i^{s}-y_i^{gt}
    \end{equation} 
    where N is the number of predicted points. 
    
    For each rarity level (i.e., normal, moderately rare, very rare, and extremely rare), we define a corresponding rarity-based penalty function $f(\cdot)$. Let $p_i$ denote the $i$-th point within the prediction window whose rarity level is, by definition, less than or equal to that of the entire predicted time series.
    
    For normal events, the rarity-based penalty function is defined as follows, which is identical to the standard function used in the MSE loss:
        \begin{IEEEeqnarray}{l}
            f_{normal}(\Delta y_i) = \Delta y_i^2
        \end{IEEEeqnarray}
        
    For moderately rare events, $f$ is defined as
        \begin{IEEEeqnarray}{l}
            f_{moderate}(\Delta y_i) = \nonumber \\
            \left\{
                \begin{array}{ll}
                     e^{-\Delta y_i}-1,& \text{if }  p_i \in \text{moderate } \text{and }  \Delta y_i \leq 0 \\
                     \Delta y_i^2,& \text{otherwise} \\
                \end{array}
            \right.
        \end{IEEEeqnarray}
which assigns equal penalties to over-predictions on moderately rare points and on points that are not moderately rare within the prediction window, while imposing a higher penalty for under-predictions.
    
    For very rare events, we reduce the penalty for over-prediction, considering that such events are significantly less frequent compared to normal and moderately rare events.
        \begin{IEEEeqnarray}{l}
        f_{very}(\Delta y_i) = \nonumber \\
            \left\{
                \begin{array}{ll}
                    \Delta y_i^2, & \text{if } p_i \notin \text{very rare } \\
                    e^{-\Delta y_i} - 1, & \text{if } p_i \in \text{very rare } \text{and } \Delta y_i \leq 0 \\
                    \log(\cosh(\Delta y_i)), & \text{if } p_i \in \text{very rare } \text{and } \Delta y_i > 0
                \end{array}
            \right.
        \end{IEEEeqnarray}

    For extremely rare events, given their very limited number of samples, we impose a strong penalty for under-prediction on extreme rare points within the prediction window, while significantly reducing the penalty for over-prediction to a level lower than that of moderately rare events. This is formally represented as follows, where $N$ denotes prediction steps:
         \begin{IEEEeqnarray}{l}
            f_{extreme}(\Delta y_i) = \nonumber \\
            \left\{
                \begin{array}{ll}
                     \Delta y_i^2,& \text{if } p_i \notin \text{extreme rare } \\
                    e^{-\Delta y_i}-1,& \text{if } p_i \in \text{extreme rare } \text{and }  \Delta y_i \leq 0 \\
                    e^{\frac{\Delta y_i}{N+1}}-1, & \text{if } p_i \in  \text{extreme rare } \text{and }  \Delta y_i > 0
                \end{array}
            \right.
        \end{IEEEeqnarray}

    This design encourages the model to be more tolerant of overestimation in rare scenarios, which is beneficial for capturing extreme patterns. Fig.~\ref{fig:h_loss} illustrates the proposed hierarchical rarity-aware penalty loss. This hierarchical loss design enables each expert model to perform more nuanced and targeted learning at its designated rarity level. It mitigates underprediction, which is particularly harmful in rare event forecasting, and encourages bolder yet controlled predictions, thereby improving model robustness under conditions of data imbalance and scarcity. Although the loss is constructed based on our defined rarity levels, it can be flexibly adapted to different rarity definitions across applications and datasets through appropriate modifications.

    The final loss function for expert model training can be represented as follows:
    \begin{equation}
        L = L_{rare} + \beta L_{KD}
    \end{equation}
    where $\beta$ is the balance parameter to control the loss from knowledge distillation. Note that knowledge distillation loss is not used for normal expert model training.

\begin{figure}
    \centering
    \includegraphics[width=0.5\linewidth]{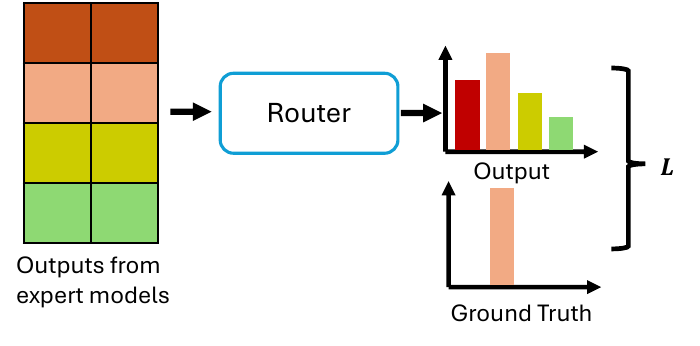}
    \vskip -1em
    \caption{The training process for router}
    \vskip -1em
    \label{fig:router}
    \vskip -1em
\end{figure}

\subsection{Router - Mixture of Expert Models}
    Although we have expert models capable of forecasting time series at different levels of rarity, the rarity level is unknown during testing, making it challenging to determine which expert should be used. A straightforward solution is to employ a classifier to select the most appropriate expert model. However, this naive approach overlooks the underlying correlations among expert models, thereby limiting the ability to leverage shared knowledge across experts to enhance rare event prediction. Thus, to further leverage the knowledge across all rarity levels and complement the prediction for rarer events, inspired by the architecture of MoE in LLMs \cite{zhou2022mixture}, we design a router to dynamically assign weights to expert models, which allows the model to effectively leverage and fuse knowledge from different rarity-level experts. We train the router based on the ranking relationships among rarity levels, ensuring that it prioritizes expert models whose rarity levels are closer to the target sample, thereby incorporating more relevant knowledge into the final prediction.


    Learning to rank is a commonly used approach for leveraging ranking information among items, typically categorized into pointwise, pairwise, and listwise methods. In pointwise ranking, items are evaluated independently; thus, one of the most commonly used objective functions is the cross-entropy loss \cite{lin2024understanding}.
    As shown in Fig. \ref{fig:router}, 
    we take the outputs from the expert models, denoted as $Y_{T:T+H}^{experts} \in \mathbb{R}^{H \times E}$, as input to the router. A MLP is then applied as a router to generate the expert weights: $\boldsymbol{\alpha} \in \mathbb{R}^E = softmax(\text{MLP}(Y_{T:T+H}^{\text{experts}}))$. Following the MoE paradigm commonly used in LLMs, the router selects the top-$k$ expert models and assigns weights accordingly
    to them. The final output is obtained as follows:
    \begin{equation}
         Y_{T:T+H} = Y_{T:T+H}^{experts} \cdot \alpha,
    \end{equation} 
    We adopt the pointwise ranking loss to train the model, which can be formulated as multi-class cross-entropy loss as follows: 
    \begin{equation}
        \mathcal{L}_{\text{router}} = -\frac{1}{N} \sum_{i=1}^{N} \log\left( \frac{\exp(\hat{y}^{(e_i)}_i)}{\sum_{e=1}^{E} \exp(\hat{y}^{(e)}_i)} \right)
    \end{equation}
   where $N$ is the number of training samples, $E$ is the number of expert models, $\hat{y}^{(e)}_i$ 
   is the logit output from the router corresponding to the $e$-th expert for the $i$-th sample, and  $e_i$  is the ground truth rarity level label for the $i$-th sample.

   This pointwise formulation is convenient and effective in our context, but alternative loss functions such as those used in pairwise or listwise ranking methods\cite{lin2024understanding} can also be explored in future work. Moreover, other MoE strategies, such as employing large time series foundation models as shared experts, also deserve further exploration.

\section{Experiment Results}

\begin{table}[t]
    \centering
    \caption{The statistics for datasets}
    \vskip -0.5em
    \begin{tabular}{l|c|c|c|c}
    \hline
       Dataset  & total points  & moderate & very & extreme \\
       \hline
       Beijing Air  & 41,775 & 2,008 (0.22) & 1,814 (0.28) & 428 (0.42) \\
       Italy Air & 914 & 43 (0.44) & 37 (0.55) & 11 (0.70) \\
       Wind Speed & 95,156 & 5,769 (0.32) & 2,897 (0.4) & 892 (0.52) \\
       Wave Height & 43,643 & 2,560 (1.97) & 1,435 (2.30) & 439 (2.82) \\
       \hline
    \end{tabular}
    \vskip -1em
    \label{tab:dataset}
        \vskip -1em
\end{table}

    In this section, we evaluate xTime on four real-world datasets to demonstrate its effectiveness. We aim to answer the following research questions: (i) \textbf{RQ1:} How good is the performance of xTime compared with state-of-the-art baselines? (ii) \textbf{RQ2:} How do the hyperparameters affect the performance of xTime? and (iii) \textbf{RQ3:} What are the contributions of different components of xTime to its overall performance?


\subsection{Experimental Setup}

\subsubsection{Datasets}We evaluate xTime on 4 real-world datasets: 2 air quality datasets from Beijing~\footnote{https://archive.ics.uci.edu/dataset/381} and Italy~\footnote{https://archive.ics.uci.edu/dataset/360}, 1 climate and weather dataset~\footnote{https://www.kaggle.com/datasets/budincsevity/szeged-weather}, and 1 oceanographic dataset~\footnote{https://www.kaggle.com/datasets/jolasa/waves-measuring-buoys-data-mooloolaba}. 
The statistics for datasets can be found in Table \ref{tab:dataset}. The values in brackets represent the thresholds used to determine the rarity levels.

\begin{table}[t]
    \caption{Experimental results for Beijing Air Quality. The best results are in \textbf{bold}, the second are marked with \underline{underline}}
    \label{tab:bjair}
    \vskip -0.5em
    \centering
    \begin{tabular}{l|c|c|c|c|c}
    \hline
    \multirow{3}{*}{Method} & \multirow{3}{*}{Metrics} & \multicolumn{4}{c}{Beijing Air Quality} \\ \cline{3-6}
    & & \multirow{2}{*}{Overall} & \multicolumn{3}{c}{Rare Points} \\ \cline{4-6}
    & & & moderate & very rare & extreme \\
    \hline
    \multirow{2}{*}{Arima} & MSE & 0.008 & 0.019 & 0.053 & 0.125 \\
    & MAE & 0.069 & 0.138 & 0.227 & 0.352 \\
    \hline
    \multirow{2}{*}{LSTM} & MSE & \underline{0.005} & 0.019 & 0.030 & 0.058 \\
    & MAE & 0.047 & 0.123 & 0.152 & 0.214 \\
    \hline
    \multirow{2}{*}{PatchTST} & MSE & 0.006 & 0.022 & \underline{0.029} & 0.057 \\
    & MAE & \underline{0.046} & 0.127 & 0.149 & 0.229 \\
    \hline\hline
    
    \multirow{2}{*}{NEC+} & MSE & 0.007 & 0.018 & 0.039 & 0.088 \\
    & MAE & 0.058 & 0.122 & 0.198 & 0.289 \\
    \hline
    \multirow{2}{*}{SaDI} & MSE & 0.006 & \underline{0.017} & 0.036 & 0.075 \\
    & MAE & 0.054 &	0.121 &	0.179 &	0.265 \\
    \hline
    \multirow{2}{*}{EPL} & MSE & 0.008 & 0.018 &	0.045 &	0.110 \\
    & MAE & 0.071 &	0.121 &	0.207 &	0.329 \\
    \hline\hline
    \multirow{2}{*}{Time-MoE} & MSE & 0.008 &	0.018 &	\underline{0.029} &	0.049 \\
    & MAE & 0.061 &	\underline{0.119} &	0.147 &	0.185 \\
    \hline
    \multirow{2}{*}{Chronos} & MSE & 0.006 &	0.02 &	0.03 &	\underline{0.034} \\
    & MAE & 0.048 &	0.123 &	\underline{0.145} &	\underline{0.162} \\
    \hline\hline
    
    \multirow{2}{*}{xTime} & MSE & \textbf{0.004} &	\textbf{0.014} &	\textbf{0.024} &	\textbf{0.027} \\
    & MAE & \textbf{0.044} &	\textbf{0.103} &	\textbf{0.136} &	\textbf{0.161} \\
    \hline
    
    \end{tabular}
    \vskip -1.5em
\end{table}

\begin{table}[t]
    \caption{Experimental results for Italy Air Quality. The best results are in \textbf{bold}, the second are marked with \underline{underline}}
    \label{tab:italyair}
    \vskip -0.5em
    \centering
    \begin{tabular}{l|c|c|c|c|c}
    \hline
    \multirow{3}{*}{Method} & \multirow{3}{*}{Metrics} & \multicolumn{4}{c}{Italy Air Quality} \\ \cline{3-6}
    & & \multirow{2}{*}{Overall} & \multicolumn{3}{c}{Rare Points} \\ \cline{4-6}
    & & & moderate & very rare & extreme \\
    \hline
    \multirow{2}{*}{Arima} & MSE & 0.042 &	0.093 &	0.188 &	0.475 \\
    & MAE & 0.167 &	0.281 &	0.373 &	0.682 \\
    \hline
    \multirow{2}{*}{LSTM} & MSE & 0.051 &	0.109 &	0.233 &	0.536 \\
    & MAE & 0.175 &	0.339 &	0.482 &	0.731 \\
    \hline
    \multirow{2}{*}{PatchTST} & MSE & 0.038 &	0.104 &	0.141 &	0.362 \\
    & MAE & 0.155 &	0.321 &	0.367 &	0.591 \\
    \hline\hline
    
    \multirow{2}{*}{NEC+} & MSE & 0.043 &	0.089 &	0.168 &	0.458 \\
    & MAE & 0.148 &	0.273 &	0.373 &	0.668 \\
    \hline
    \multirow{2}{*}{SaDI} & MSE & 0.041 &	0.088 &	0.179 &	0.461 \\
    & MAE & \underline{0.134} &	0.287 &	0.379 &	0.679 \\
    \hline
    \multirow{2}{*}{EPL} & MSE & 0.046 &	0.099 &	0.177 &	0.422 \\
    & MAE & 0.174 &	0.269 &	0.419 &	0.649 \\
    \hline\hline
    
    \multirow{2}{*}{Time-MoE} & MSE & 0.042 &	\underline{0.084}& 0.139&	0.488 \\
    & MAE & 0.151 &	0.269	&0.343	&0.686 \\
    \hline
    \multirow{2}{*}{Chronos} & MSE & \underline{0.03}	&0.086	&\underline{0.135}	& \underline{0.358} \\
    & MAE & 0.138&	\underline{0.267}	& \underline{0.336}	& \underline{0.588} \\
    \hline\hline
    
    \multirow{2}{*}{xTime} & MSE & \textbf{0.027}&	\textbf{0.08}&	\textbf{0.127}&	\textbf{0.324} \\
    & MAE & \textbf{0.129}&	\textbf{0.246}&	\textbf{0.335}&	\textbf{0.572} \\
    \hline
    
    \end{tabular}
    \vskip -1.5em
\end{table}

\subsubsection{Implementation} We split each dataset into training, validation, and test sets in an 8:1:1 ratio. The thresholds for extreme, very rare, and moderately rare levels can be found in Table \ref{tab:dataset}, respectively. We focus on the overall performance within the prediction window as well as the performance on rare points.
Following the experimental setting in prior work~\cite{wang2024self}, we set $B=4$. For Beijing Air and Wind Speed, the history window is 128 and the prediction length is 24. For Italy Air, they are 24 and 8, respectively. For Wave Height, we use 512 and 96 as history and prediction length. Evaluations are conducted on normalized data except for Wave Height, which uses raw values. For rare point evaluation, we compute MSE and MAE for each rarity level: $MSE = \frac{1}{M}\sum_{i=1}^{M}(y_i^{truth}-y_i^{c})^2$ and $MAE = \frac{1}{M}\sum_{i=1}^{M}|y_i^{truth}-y_i^{c}|$, where $M$ denotes the number of points belonging to the respective rarity level..

\begin{table}[t]
    \caption{Experimental results for wind speed. The best results are in \textbf{bold}, the second are marked with \underline{underline}}
    \label{tab:ws}
    \vskip -0.5em
    \centering
    \begin{tabular}{l|c|c|c|c|c}
    \hline
    \multirow{3}{*}{Method} & \multirow{3}{*}{Metrics} & \multicolumn{4}{c}{Wind Speed} \\ \cline{3-6}
    & & \multirow{2}{*}{Overall} & \multicolumn{3}{c}{Rare Points} \\ \cline{4-6}
    & & & moderate & very rare & extreme \\
    \hline
    \multirow{2}{*}{Arima} & MSE &0.009 &  0.030 &  0.069 &  0.143 \\
    & MAE & 0.073 &  0.172 &  0.262 &  0.376 \\
    \hline
    \multirow{2}{*}{LSTM} & MSE & \underline{0.007} &  0.030 &  0.059 &  0.132 \\
    & MAE & 0.062 &  0.158 &  0.238 &  0.352  \\
    \hline
    \multirow{2}{*}{PatchTST} & MSE & \underline{0.007} &  0.027 &  0.057 &  0.129 \\
    & MAE & \underline{0.060} &  0.152 &  0.226 &  0.348 \\
    \hline\hline
    
    \multirow{2}{*}{NEC+} & MSE & 0.008 &  0.028 &  0.063 &  0.143 \\
    & MAE & 0.069 &  0.161 &  0.247 &  0.378 \\
    \hline
    \multirow{2}{*}{SaDI} & MSE & 0.008 &  \underline{0.025} &  0.058 &  0.128 \\
    & MAE & 0.066 &  0.153 &  0.234 &  0.350 \\
    \hline
    \multirow{2}{*}{EPL} & MSE & 0.009 &  0.032 &  0.072 &  0.139 \\
    & MAE & 0.074 &  0.177 &  0.265 &  0.371 \\
    \hline\hline
    \multirow{2}{*}{Time-MoE} & MSE & 0.010 &  0.028 &  0.055 &  0.132  \\
    & MAE & 0.073 &  0.149 &  0.214 &  0.341 \\
    \hline
    \multirow{2}{*}{Chronos} & MSE & 0.009 &  0.026 &  \underline{0.051} &  \underline{0.128} \\
    & MAE & 0.067 &  \underline{0.142} &  \underline{0.206} &  \underline{0.339}  \\
    \hline\hline
    
    \multirow{2}{*}{xTime} & MSE & \textbf{0.006} &  \textbf{0.016} &  \textbf{0.043} &  \textbf{0.104}  \\
    & MAE & \textbf{0.058} &  \textbf{0.115} &  \textbf{0.171} &  \textbf{0.303}\\
    \hline
    
    \end{tabular}
        \vskip -1.5em
\end{table}

\begin{table}[t]
    \caption{Experimental results for wave height. The best results are in \textbf{bold}, the second are marked with \underline{underline}}
    \label{tab:wh}
    \vskip -0.5em
    \centering
    \begin{tabular}{l|c|c|c|c|c}
    \hline
    \multirow{3}{*}{Method} & \multirow{3}{*}{Metrics} & \multicolumn{4}{c}{Wave Height} \\ \cline{3-6}
    & & \multirow{2}{*}{Overall} & \multicolumn{3}{c}{Rare Points} \\ \cline{4-6}
    & & & moderate & very rare & extreme \\
    \hline
    \multirow{2}{*}{Arima} & MSE & 0.263 &  0.458 &  0.983 &  2.225 \\
    & MAE &0.420 &  0.642 &  0.989 &  1.491 \\
    \hline
    \multirow{2}{*}{LSTM} & MSE & 0.195 &  0.485 &  0.748 &  1.163 \\
    & MAE & 0.338 &  0.629 &  0.756 &  1.055 \\
    \hline
    \multirow{2}{*}{PatchTST} & MSE & 0.144 &  0.528 &  0.804 &  0.826 \\
    & MAE & 0.278 &  0.660 &  0.747 &  0.875  \\
    \hline\hline
    
    \multirow{2}{*}{NEC+} & MSE & \underline{0.135} &  0.460 &  0.688 &  1.041 \\
    & MAE & 0.309 &  0.672 &  0.791 &  1.003 \\
    \hline
    \multirow{2}{*}{SaDI} & MSE & 0.138 &  \underline{0.457} &  0.696 &  1.122 \\
    & MAE & 0.304 &  0.644 &  0.787 &  1.042 \\
    \hline
    \multirow{2}{*}{EPL} & MSE & 0.140 &  0.464 &  0.613 &  1.760 \\
    & MAE & 0.312 &  0.741 &  0.775 &  1.320 \\
    \hline\hline
    \multirow{2}{*}{Time-MoE} & MSE & 0.210 &  0.892 &  1.323 &  1.770 \\
    & MAE & 0.301 &  0.848 &  0.975 &  1.266 \\
    \hline
    \multirow{2}{*}{Chronos} & MSE & 0.144 &  0.513 &  \underline{0.565} &  \underline{0.618} \\
    & MAE & \underline{0.271} &  \underline{0.618} & \underline{0.617} &  \underline{0.787}  \\
    \hline\hline
    
    \multirow{2}{*}{xTime} & MSE & \textbf{0.088} &  \textbf{0.448} &  \textbf{0.510} &  \textbf{0.596} \\
    & MAE & \textbf{0.223} &  \textbf{0.591} &  \textbf{0.557} &  \textbf{0.720} \\
    \hline
    
    \end{tabular}
        \vskip -1.5em
\end{table}

\subsubsection{Baselines} We compare xTime with 8 state-of-the-art baselines, including 3 well-known time series models: Arima~\cite{box2015time}, LSTM~\cite{hochreiter1997long}, and PatchTST~\cite{Yuqietal-2023-PatchTST}; 3 models specified for extreme events: NEC+~\cite{li2023extreme}, SaDI~\cite{Liu2023Sadi}, and EPL~\cite{wang2024self}; and 2 large time series models: Time-MoE \cite{shi2024time}, Chronos~\cite{ansari2024chronos}.
\begin{itemize}[leftmargin=*]
    \item \textbf{Arima:} A classical statistical model for time series analysis and forecasting, well-suited for linear patterns in data.
    \item \textbf{LSTM:} A type of recurrent neural network (RNN) specifically designed to model and learn from sequential dependencies in time series data.
    \item \textbf{PatchTST:} A Transformer-based model that transforms multivariate time series into univariate time series and leverages patching techniques to aggregate temporal information within fixed-length patches.
    \item \textbf{NEC+:} A model designed for extreme event prediction that first classifies events into extreme and normal categories and then applies specialized models for each category.
    \item \textbf{SaDI:} A feature-engineering-based method for extreme event forecasting that extracts discriminative features from time series data to improve prediction.
    \item \textbf{EPL:} A method that introduces an extreme penalty loss specifically tailored for decomposed time series, designed to apply targeted rewards or penalties to predictions involving extreme events.
    \item \textbf{Time-MoE:} A large-scale time series forecasting model that incorporates the MoE paradigm to enhance model specialization and scalability.
    \item \textbf{Chronos:} A cutting-edge time series model that performs forecasting via next-token prediction, effectively modeling complex temporal dynamics.
\end{itemize}

\begin{table}[t]
    \caption{Experimental results for Beijing Air Quality with different numbers of expert models. The best results are in \textbf{bold}, the second are marked with \underline{underline}}
    \label{tab:abl_bjair}
    \vskip -0.5em
    \centering
    \begin{tabular}{l|c|c|c|c|c}
    \hline
    \multirow{3}{*}{Method} & \multirow{3}{*}{Metrics} & \multicolumn{4}{c}{Beijing Air Quality} \\ \cline{3-6}
    & & \multirow{2}{*}{Overall} & \multicolumn{3}{c}{Rare Points} \\ \cline{4-6}
    & & & moderate & very rare & extreme \\
    \hline

    \multirow{2}{*}{xTime (k=4)} & MSE & \underline{0.005} &	0.015 &	0.027 &	0.033 \\
    & MAE & \underline{0.045} &	0.109 &	0.142 &	0.166 \\
    \hline
    \multirow{2}{*}{xTime (k=3)} & MSE & \textbf{0.004} &	\underline{0.014} &	0.026 &	\underline{0.032} \\
    & MAE & 0.046 &	0.107 &	0.14 &	0.164 \\
    \hline
    \multirow{2}{*}{xTime (k=2)} & MSE & \textbf{0.004} &	\underline{0.014} &	\underline{0.024} &	\textbf{0.027} \\
    & MAE & \textbf{0.044} &	\underline{0.103} &	\underline{0.136} &	\underline{0.161} \\
    \hline
    \multirow{2}{*}{xTime (k=1)} & MSE & \textbf{0.004} &	\textbf{0.012} &	\textbf{0.022} &	\textbf{0.027} \\
    & MAE & \underline{0.045} &	\textbf{0.095} &	\textbf{0.124} &	\textbf{0.143} \\
    \hline
    
    \end{tabular}
    \vskip -1.5em
\end{table}

\begin{table}[t]
    \caption{Experimental results for Italy Air Quality with different numbers of expert models. The best results are in \textbf{bold}, the second are marked with \underline{underline}}
    \label{tab:abl_italyair}
    \vskip -0.5em
    \centering
    \begin{tabular}{l|c|c|c|c|c}
    \hline
    \multirow{3}{*}{Method} & \multirow{3}{*}{Metrics} & \multicolumn{4}{c}{Italy Air Quality} \\ \cline{3-6}
    & & \multirow{2}{*}{Overall} & \multicolumn{3}{c}{Rare Points} \\ \cline{4-6}
    & & & moderate & very rare & extreme \\
    \hline
    
    \multirow{2}{*}{xTime (k=4)} & MSE & \textbf{0.026}&	\textbf{0.077}&	0.129&	0.35 \\
    & MAE & \textbf{0.127}&	\textbf{0.242}&	0.338&	0.584 \\
    \hline
    \multirow{2}{*}{xTime (k=3)} & MSE & \underline{0.027}&	\underline{0.079}&	\underline{0.128}&	\underline{0.325} \\
    & MAE & \underline{0.128}&	\underline{0.243}&	\textbf{0.332}&	\underline{0.574} \\
    \hline
    \multirow{2}{*}{xTime (k=2)} & MSE & \underline{0.027}&	0.08&	\textbf{0.127}&	\textbf{0.324} \\
    & MAE & 0.129&	0.246&	\underline{0.335}&	\textbf{0.572} \\
    \hline
    \multirow{2}{*}{xTime (k=1)} & MSE & 0.028&	0.081&	0.135&	0.336 \\
    & MAE & 0.133&	0.252	&0.337	&0.578 \\
    \hline
    
    \end{tabular}
    \vskip -1.5em
\end{table}

\begin{table}[t]
    \caption{Experimental results for wind speed with different numbers of expert models. The best results are in \textbf{bold}, the second are marked with \underline{underline}}
    \label{tab:abl_ws}
    \vskip -0.5em
    \centering
    \begin{tabular}{l|c|c|c|c|c}
    \hline
    \multirow{3}{*}{Method} & \multirow{3}{*}{Metrics} & \multicolumn{4}{c}{Wind Speed} \\ \cline{3-6}
    & & \multirow{2}{*}{Overall} & \multicolumn{3}{c}{Rare Points} \\ \cline{4-6}
    & & & moderate & very rare & extreme \\
    \hline

    \multirow{2}{*}{xTime (k=4)} & MSE & 0.007 &  0.017 &  0.044 &  0.107  \\
    & MAE &  0.059 &  0.116 &  0.174 &  0.312\\
    \hline
    \multirow{2}{*}{xTime (k=3)} & MSE &  \underline{0.006} & \underline{ 0.016} &  0.044 &  0.107  \\
    & MAE &  \underline{0.058} &  \underline{0.115} &  0.172 &  0.310 \\
    \hline
    \multirow{2}{*}{xTime (k=2)} & MSE & \underline{0.006} &  \underline{ 0.016} &  \underline{0.043} &  \underline{0.104}  \\
    & MAE & \underline{0.058} &  \underline{0.115} &  \underline{0.171} &  \underline{0.303}\\
    \hline
    \multirow{2}{*}{xTime (k=1)} & MSE & \textbf{0.005} &  \textbf{0.012} &  \textbf{0.035} &  \textbf{0.094}  \\
    & MAE & \textbf{0.057} &  \textbf{0.096} &  \textbf{0.169} &  \textbf{0.287}  \\
    \hline
    
    \end{tabular}
    \vskip -1.5em
\end{table}

\begin{table}[t]
    \caption{Experimental results for wave height with different numbers of expert models. The best results are in \textbf{bold}, the second are marked with \underline{underline}}
    \label{tab:abl_wh}
    \vskip -0.5em
    \centering
    \begin{tabular}{l|c|c|c|c|c}
    \hline
    \multirow{3}{*}{Method} & \multirow{3}{*}{Metrics} & \multicolumn{4}{c}{Wave Height} \\ \cline{3-6}
    & & \multirow{2}{*}{Overall} & \multicolumn{3}{c}{Rare Points} \\ \cline{4-6}
    & & & moderate & very rare & extreme \\
    \hline
    
    \multirow{2}{*}{xTime (k=4)} & MSE & 0.095 &  \textbf{0.441} &  0.511 &  0.636 \\
    & MAE & 0.238 &  \textbf{0.584} &  0.573 &  0.748 \\
    \hline
    \multirow{2}{*}{xTime (k=3)} & MSE & 0.092 &  0.450 &  \textbf{0.507} &  \underline{0.614} \\
    & MAE & 0.233 &  \underline{0.586} &  \underline{0.566} &  0.735 \\
    \hline
    \multirow{2}{*}{xTime (k=2)} & MSE & \underline{0.088} &  \underline{0.448} &  \underline{0.510} &  \textbf{0.596} \\
    & MAE & \underline{0.223} &  0.591 &  \textbf{0.557} &  \underline{0.720} \\
    \hline
    \multirow{2}{*}{xTime (k=1)} & MSE & \textbf{0.085} &  0.453 &  0.559 &  0.616  \\
    & MAE & \textbf{0.209} &  0.611 &  0.582 &  \textbf{0.716} \\
    \hline
    
    \end{tabular}
    \vskip -1.5em
\end{table}

\subsection{Performance Comparison With Baselines}
We conduct comprehensive experiments to evaluate the performance of xTime against eight state-of-the-art baseline models under consistent settings, aiming to answer RQ1. Tables \ref{tab:bjair}, \ref{tab:italyair}, \ref{tab:ws}, and \ref{tab:wh} present the evaluation results across all baselines and xTime.

Among the baselines, PatchTST demonstrates strong performance on rare event prediction due to its patching strategy, which effectively captures richer temporal patterns from the time series data. Additionally, large time series models benefit from extensive training data and exhibit relatively better performance, particularly on rare points.

However, compared with all these methods, our proposed model, xTime, achieves superior performance on both overall metrics and rare event evaluations. Specifically, compared to PatchTST, xTime shows MSE improvements on rare points in the following ranges: $[17\%, 52\%]$ for the Beijing air quality dataset, $[9\%, 23\%]$ for the Italy air dataset, $[17\%, 40\%]$ for the wind speed dataset, and $[15\%, 36\%]$ for the wave height dataset. When compared with Chronos, xTime achieves MSE improvements of $[20\%, 30\%]$ on the Beijing air dataset, $[6\%, 9\%]$ on the Italy air dataset, $[17\%, 38\%]$ on the wind speed dataset, and $[3\%, 13\%]$ on the wave height dataset.
Notably, xTime attains these results while utilizing less training data than large time series models, yet still matches or even surpasses their performance. These findings reaffirm the effectiveness of xTime in extreme event prediction.

\subsection{Hyper-parameter Sensitivity Analysis}
In this section, we evaluate two hyperparameters, the balance control parameter $\beta$ and the number of experts $k$, in xTime to answer \textbf{RQ2}.

\begin{figure*}[t]
    \centering

    \subfigure[Italy air dataset]{\includegraphics[trim = {15 0 0 0}, clip, width = 0.48\linewidth]{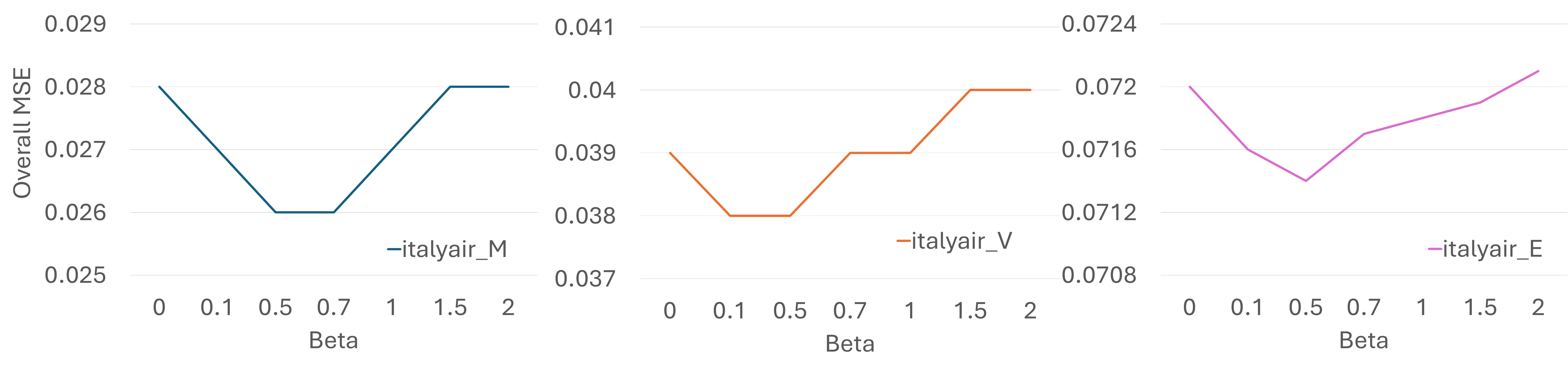} \label{subfig:italyair}} \quad
    \subfigure[Beijing air dataset]{\includegraphics[trim = {15 0 0 0}, clip,width = 0.48\linewidth]{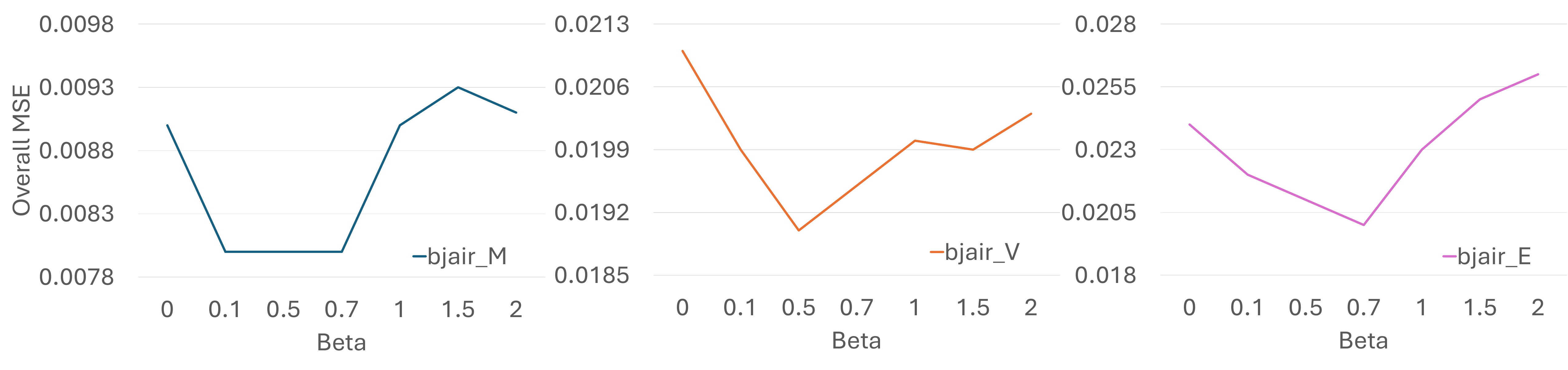} \label{subfig:bjair}} \\ 
    \subfigure[Wind Speed dataset]{\includegraphics[trim = {15 0 0 0}, clip,width = 0.48\linewidth]{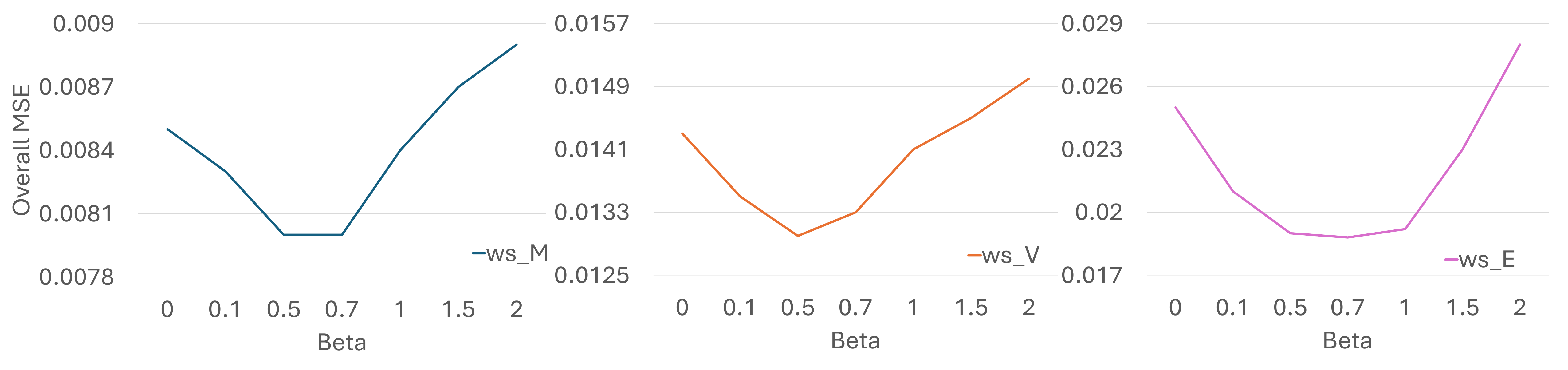} \label{subfig:ws}}\quad
    \subfigure[Wave height dataset]{\includegraphics[trim = {15 0 0 0}, clip,width = 0.48\linewidth]{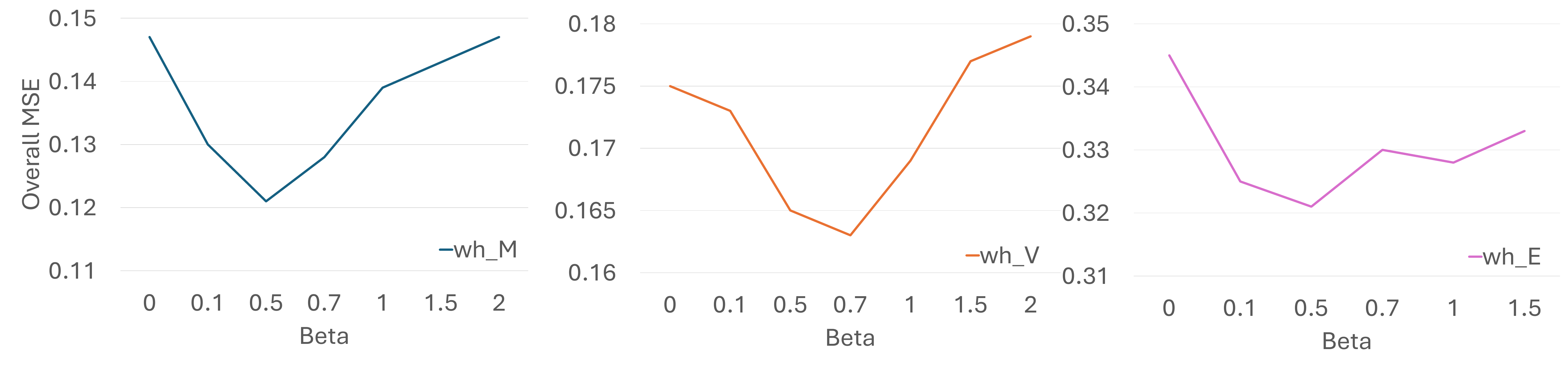} \label{subfig:wh}}\\ 
    \vskip -0.5em
    \caption{Hyper-parameters $\beta$ evaluation on different dataset: (a) Italy Air dataset (b) Beijing air dataset (c) Wind speed dataset (d) Wave height dataset. The letters M, V, and E in the figure denote the moderate rare mode, very rare model, and the extremely rare model, respectively.}
    \vspace{-1.5em}
    \label{fig:hyper}
\end{figure*}

\noindent\textbf{Balance Parameter $\beta$:} In xTime, we use a balance parameter $\beta$ to control the knowledge distillation loss and ground truth loss when training the rare event prediction models. We set $\beta = \{0, 0.1, 0.5, 0.7, 1, 1.5, 2\}$ and conducted the evaluation. From Fig. \ref{fig:hyper}, we observe that as the value of $\beta$ increases, the expert models initially show improved performance, reaching an optimal range between $0.5$ and $0.7$. However, further increasing $\beta$ leads to a decline in performance. This is expected, as an excessively high $\beta$ causes the expert models to rely too heavily on the teacher model, thereby neglecting valuable information from the ground truth. As a result, the models struggle to effectively learn the characteristics of rare events within their designated rarity levels.

\begin{figure}[t]
    \centering
    \includegraphics[trim= {100 0 80 0}, clip, width=\linewidth]{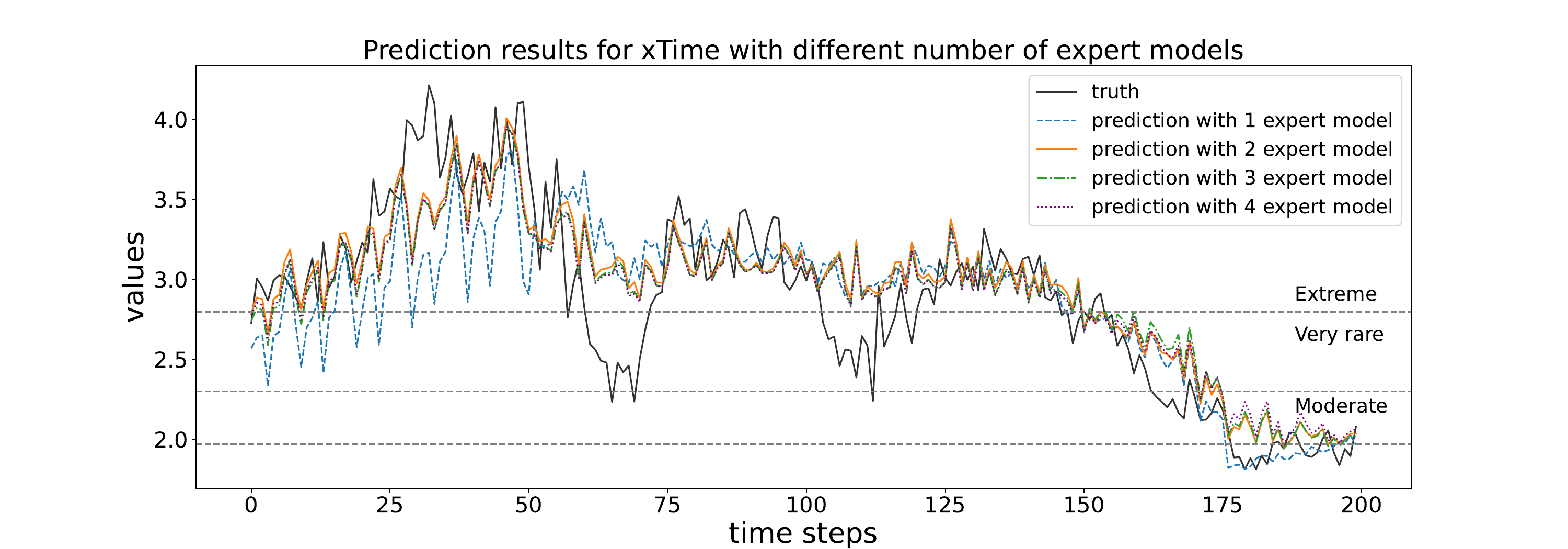}
    \vskip -1em
    \caption{Visualization for wave height with different numbers of expert models 
    }
    \vskip -1.5em
    \label{fig:differehtk}
\end{figure}

\noindent\textbf{Number of Expert Models $k$:}
Tables \ref{tab:abl_bjair}, \ref{tab:abl_italyair}, \ref{tab:abl_ws}, and \ref{tab:abl_wh} present the evaluation results of xTime with varying numbers of expert models ($k$) selected by the router. While xTime consistently outperforms all state-of-the-art baselines across different values of $k$, its performance varies with the number of experts. Notably, configurations with $k \in [1, 3]$ yield better results, with $k = 2$ offering the most balanced performance across both overall and rare event evaluations. This suggests that the router effectively selects the most relevant experts to enhance prediction. In contrast, involving too many expert models may introduce noise and irrelevant information, potentially degrading performance. Fig. \ref{fig:differehtk} presents the wave height forecasting results for different values of $k$. We observe that, on the wave height dataset, setting $k=2$ yields the best performance on extreme events, followed by $k=1$. These observations align with the results shown in Table \ref{tab:abl_wh}.
For the Beijing air, wind speed, and wave height datasets, xTime achieves relatively better performance with $k=1$ or $k=2$. These datasets contain a larger proportion of normal data, enabling the expert models to be effectively trained via knowledge distillation from intermediate events. Consequently, the models can also handle corresponding rare events effectively. When the router accurately assigns weights, the final predictions benefit accordingly. In contrast, for the Italy air dataset, which has fewer data points, performance improves with a larger number of expert models. This is because xTime uses its router to draw on information from all expert models, which helps improve performance when the available data is limited.




\begin{table}[t]
    \caption{Ablation study: Wavelet Transform (WT), Hierarchical Rare Penalty Loss (RP), and Knowledge Distillation (KD)}
    \label{tab:ablation}
    \vskip -0.5em
    \centering
    \tabcolsep=4pt
    \begin{tabular}{|c|c|c||c|c|c|c|}
    \hline
         WT & RP & KD & Beijing Air & Wind Speed & Italy Air & Wave Height \\
         \hline
         & & & 0.006 & 0.007 & 0.038 & 0.144\\
         \hline
         \checkmark &  &  & 0.0058 & 0.0072 & 0.028 & 0.107 \\
         \hline
         \checkmark & & \checkmark & 0.0053 & 0.007 & 0.027 & 0.095 \\
         \hline
         \checkmark & \checkmark & & 0.005 & 0.0069 & 0.027 & 0.092 \\
         \hline
         \checkmark & \checkmark & \checkmark & 0.004 & 0.006 & 0.026 & 0.088 \\
         \hline
    \end{tabular}
    \vskip -1.5em
\end{table}

\subsection{Ablation Study}

In this section, we conduct an ablation study to evaluate the contribution of each component in xTime and address \textbf{RQ3}. Specifically, xTime is composed of three core components: (1) wavelet transform (WT); (2) hierarchical rare penalty loss (RP); and (3) knowledge distillation (KD). The router, which is responsible for fusing knowledge from multiple expert models, is not evaluated independently in this study, as its effectiveness has already been validated in experiments with varying numbers of expert models, and it is essential for producing the final output. Therefore, we fix the number of selected expert models to $k=2$ and use the router to perform knowledge fusion consistently across all ablation settings.

Table \ref{tab:ablation} presents the results of the ablation study conducted on four datasets. We progressively add each component and evaluate the model using the overall MSE. The results show that all components contribute positively to the performance of xTime.The wavelet transform improves the model’s ability to capture sharp transitions in time series, while the hierarchical rare penalty loss and knowledge distillation provide additional gains by addressing data imbalance and scarcity. The best performance is achieved when both components are integrated, highlighting their complementary strengths. This ablation study reaffirms the effectiveness of each component and emphasizes their individual and combined contributions to the overall model performance.




\section{Conclusion}
In this paper, we address the challenge of extreme event prediction and propose xTime, a novel framework that integrates knowledge distillation and the MoE paradigm to enhance predictive performance. xTime leverages knowledge distillation to transfer insights from common and intermediate events, thereby mitigating the impact of data imbalance typically encountered in extreme event scenarios. To further strengthen learning, we introduce a hierarchical rare penalty loss that exploits the ordinal nature of rarity levels, encouraging each expert model to specialize in its corresponding degree of rarity. Additionally, we design a router module, inspired by the MoE mechanism in large language models, to effectively fuse the outputs from multiple expert models based on their relevance to the current input. Extensive experiments across four real-world datasets demonstrate that xTime consistently outperforms state-of-the-art baselines, highlighting its effectiveness, robustness, and practical significance for extreme event prediction tasks.

\section{Acknowledgment}
This work is supported by gift funding from NEC Labs.


\bibliographystyle{ieeetr}
\bibliography{reference}

\end{document}